\documentclass{article} % For LaTeX2e

\usepackage{iclr2026_conference,times}
\iclrfinalcopy

\usepackage{fancyhdr}
\usepackage{xcolor} % for the continuation symbol (optional)

\usepackage[utf8]{inputenc}   % allows UTF-8 encoded characters
\usepackage[T1]{fontenc}      % proper encoding for accented characters
\usepackage{amsmath,amsfonts,bm}

\def\eqref#1{equation~\ref{#1}}
\def\1{\bm{1}}

\DeclareMathAlphabet{\mathsfit}{\encodingdefault}{\sfdefault}{m}{sl}
\SetMathAlphabet{\mathsfit}{bold}{\encodingdefault}{\sfdefault}{bx}{n}

\newcommand{\E}{\mathbb{E}}

\DeclareMathOperator*{\argmax}{arg\,max}

\usepackage{hyperref}
\usepackage{url}
\usepackage{caption}
\usepackage{graphicx}
\usepackage{multirow}
\usepackage{xcolor}
\usepackage{amsmath}
\usepackage{amsthm}
\usepackage{amssymb}
\usepackage{amsfonts}
\usepackage{stmaryrd}

\usepackage{tabularx}
\usepackage{booktabs}
\usepackage{algorithm}
\usepackage{algorithmic}
\usepackage{amsthm}
\usepackage{enumitem}

\usepackage{xcolor} % in your preamble
\usepackage[table,xcdraw]{xcolor}
\usepackage{colortbl}
\usepackage{arydshln}
\usepackage{booktabs}

\newtheorem{theorem}{Theorem}

\newtheorem{assumption}{Assumption}
\usepackage[most]{tcolorbox}
\tcolorboxenvironment{theorem}{
  enhanced,
  colback=gray!5,      % light background
  colframe=black,      % border color
  coltitle=black,      % title color
  fonttitle=\bfseries, % bold title
  boxrule=0.8pt,       % border thickness
  arc=3pt,             % rounded corners
  left=6pt,right=6pt,top=6pt,bottom=6pt
}
\tcolorboxenvironment{assumption}{
  enhanced,
  colback=gray!2,      % light background
  colframe=black,      % border color
  coltitle=black,      % title color
  fonttitle=\bfseries, % bold title
  boxrule=0.8pt,       % border thickness
  arc=3pt,             % rounded corners
  left=6pt,right=6pt,top=6pt,bottom=6pt
}
\newcolumntype{Y}{>{\centering\arraybackslash}X}

\definecolor{candypink}{rgb}{1.0, 0.41, 0.71}
\definecolor{metablue}{rgb}{0.2, 0.4, 0.8}
\definecolor{phthaloblue}{rgb}{0.0, 0.15, 0.45}
\definecolor{green(ncs)}{rgb}{0.0, 0.62, 0.42}
\definecolor{lightwheat}{rgb}{1.0, 0.98, 0.82}
\definecolor{rosepink}{HTML}{F4A7B9}
\definecolor{skyblue}{HTML}{A7C7E7}

\usepackage{listings}
\usepackage{xcolor}

\definecolor{diffstart}{rgb}{0.0,0.0,0.6}
\definecolor{diffincl}{rgb}{0.0,0.6,0.0}
\definecolor{diffrem}{rgb}{0.6,0.0,0.0}

\lstdefinelanguage{diff}{
  morecomment=[f][\color{diffstart}]{@@},     % hunk headers
  morecomment=[f][\color{diffincl}]{+},      % added lines
  morecomment=[f][\color{diffrem}]{-},       % removed lines
  morecomment=[f][\color{black}]{diff},      % diff start
}

\lstdefinestyle{diffstyle}{
  language=diff,
  basicstyle=\ttfamily\small,
  showstringspaces=false,
  breaklines=true,
  postbreak=\mbox{\textcolor{red}{$\hookrightarrow$}\space},
}

\title{Huxley-G\"odel Machine: Human-Level Coding Agent Development by an Approximation of the Optimal Self-Improving Machine}

\author{Wenyi Wang\thanks{equal contribution} \quad
Piotr Pi\k{e}kos\textsuperscript{*} \quad
Li Nanbo \quad
Firas Laakom \quad
\textbf{Yimeng Chen} \\
\textbf{Mateusz Ostaszewski} \quad
\textbf{Mingchen Zhuge} \quad
\textbf{J\"{u}rgen Schmidhuber} \\
\texttt{\small\{wenyi.wang, piotr.piekos, nanbo.li, firas.laakom, yimeng.chen},\\ \texttt{\small mateusz.ostaszewski, mingchen.zhuge, juergen.schmidhuber\}@kaust.edu.sa}
\\
King Abdullah University of Science and Technology (KAUST) \\
Thuwal, Saudi Arabia
}

\begin{document}

\maketitle

\fancyhead{}

\begin{abstract}
Recent studies operationalize self-improvement through coding agents that edit their own codebases. They grow a tree of self-modifications through expansion strategies that favor higher software engineering benchmark performance, assuming that this implies more promising subsequent self-modifications.
However, we identify a mismatch between the agent's self-improvement potential (metaproductivity) and its coding benchmark performance, namely the \emph{Metaproductivity-Performance~Mismatch}. 
Inspired by Huxley’s concept of clade, we propose a metric ($\mathrm{CMP}$) that aggregates the benchmark performances of the \emph{descendants} of an agent as an indicator of its potential for self-improvement.
We show that, in our self-improving coding agent development setting, access to the true $\mathrm{CMP}$ is sufficient to simulate how the G\"odel Machine would behave under certain assumptions.
We introduce the Huxley-G\"odel Machine (HGM), which, by estimating $\mathrm{CMP}$ and using it as guidance, searches the tree of self-modifications.
On SWE-bench Verified and Polyglot, HGM outperforms prior self-improving coding agent development methods while using fewer allocated CPU hours. 
Last but not least, HGM demonstrates strong transfer to other coding datasets and LLMs. %large language models. 
The agent optimized by HGM on SWE-bench Verified with GPT-5-mini and evaluated on SWE-bench Lite with GPT-5 \textbf{achieves human-level performance, matching the best officially checked results of human-engineered coding agents}. Our code is publicly available at \url{https://github.com/metauto-ai/HGM}.

\end{abstract}

\begin{figure}[hb!]
    \centering
    \includegraphics[width=0.48\linewidth]{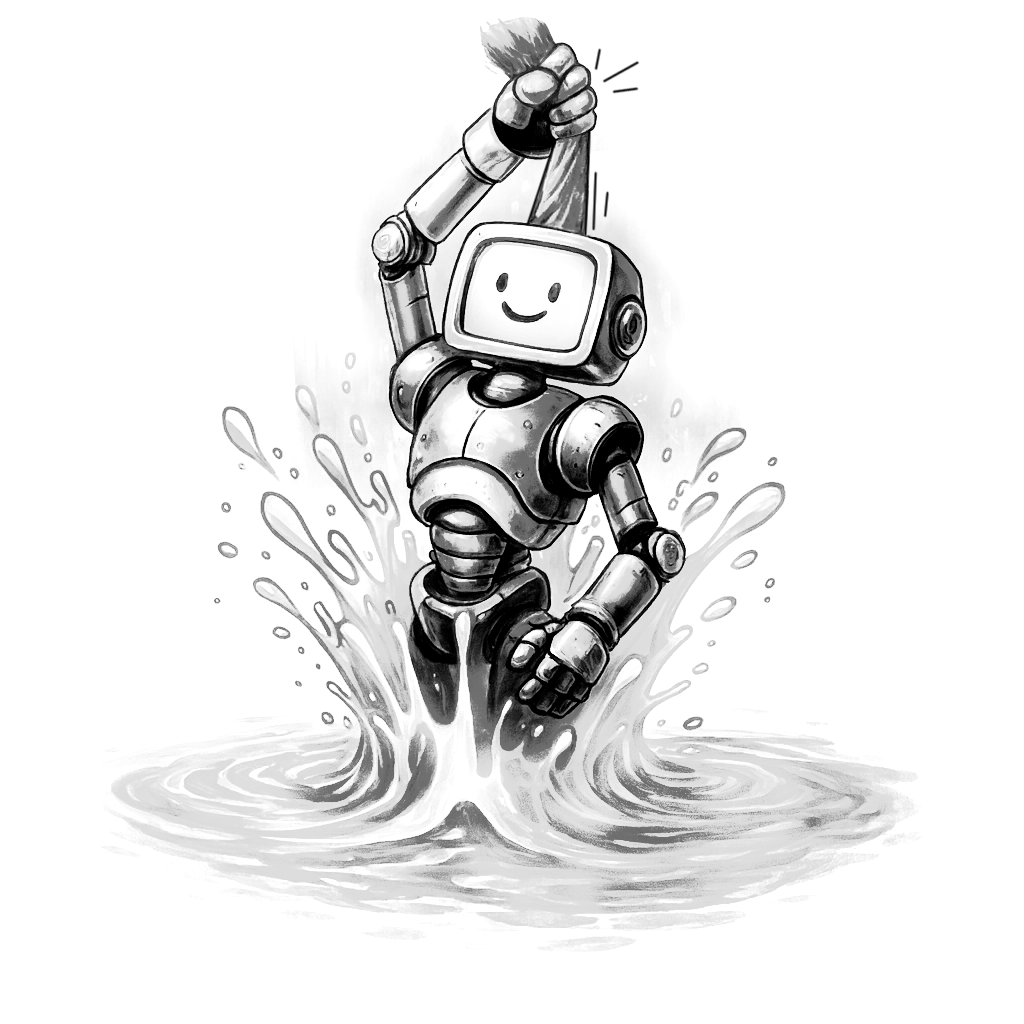}
\end{figure}

\begin{figure}[hb!]

    \centering
    \includegraphics[width=0.96\linewidth]{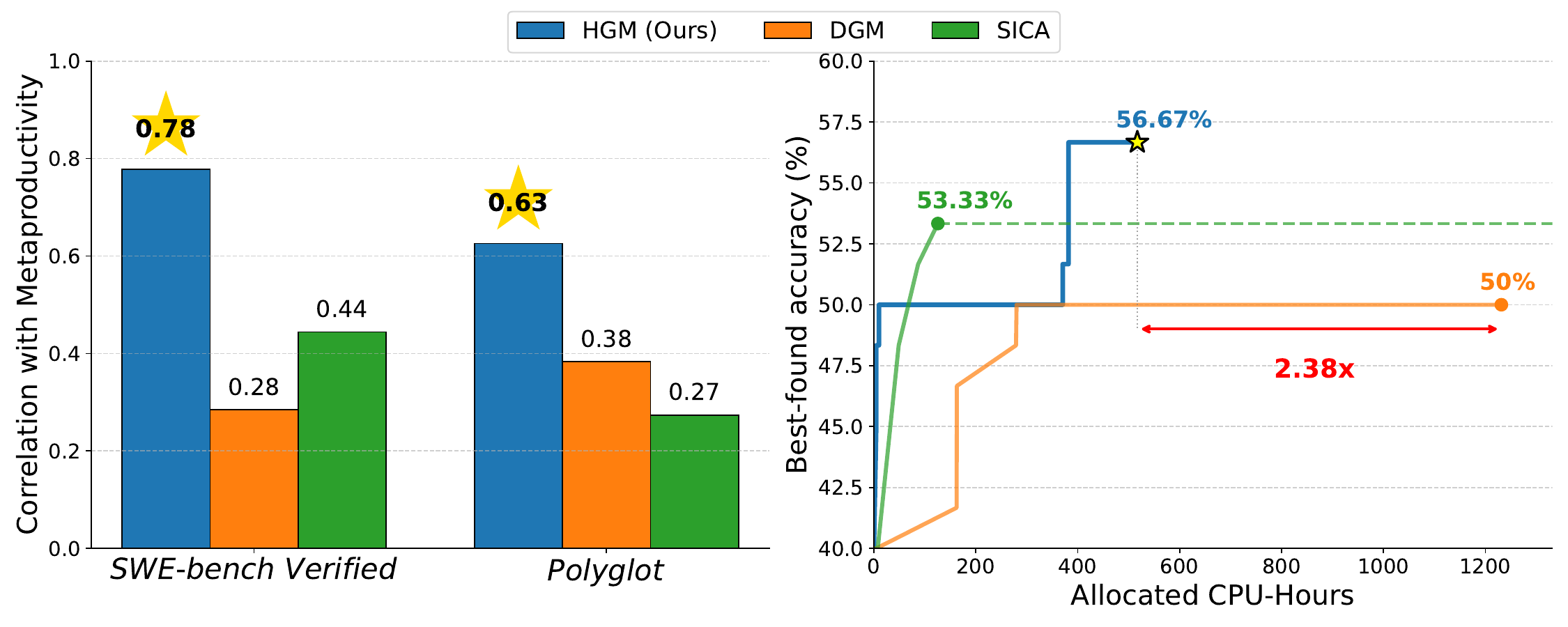}
    \caption{ \textbf{(Left)} Weak correlation between the guidance metrics of other methods (based on benchmark performance) and long-term self improvement; HGM mitigates this mismatch by leveraging clade-level metaproductivity. 
    \textbf{(Right)} On \textit{SWE-bench Verified}, HGM achieves \textbf{higher} accuracy with $2.38$ time less allocated CPU-hours. Together, the results indicate the practical advantage of approximating G\"odel Machines with long-term self-improvement estimates. Note that SICA encountered repeated errors after consuming 45\% of its budget, preventing any further self-modifications.}
    \label{fig:intro}
\end{figure}

\section{Introduction}

Processes of self-modification drive the growth of complex systems, from biological evolution~\citep{hendrikse2007evolvability,dawkins2019evolution} to cultural and scientific innovation~\citep{good1966speculations,hall2007self}.
These general ideas have been instantiated in concrete algorithms for self-improving agents~\citep{Schmidhuber1987EvolutionaryPI,Schmidhuber:03gm,nivel2013bounded,everitt2016self}, demonstrating how abstract principles of self-modification can be translated into operational mechanisms. Unlike static systems constrained by fixed architectures, such agents can incrementally modify their own self-modification mechanisms and learning strategies, reusing newly gained abilities to fuel subsequent improvements. This capacity fosters continual adaptation, reduces reliance on human intervention, and enables problem-solving capabilities that cannot be fully anticipated at design time.

A central challenge is how to decide which self-modifications to accept. The G\"odel machine~\citep{Schmidhuber:03gm} (GM) offers a theoretically optimal answer: accept only modifications that provably increase the expected long-term utility. While this provides a sound blueprint, its reliance on formal proofs makes it practically challenging. Recent implementations instead rely on coding agents that edit their own codebases and favor self-modifications from agents with higher benchmark performance~\citep{robeyns2025selfimprovingcodingagent,zhang2025dgm}. Yet, as illustrated in Figure~\ref{fig:intro} (left), this heuristic can be misleading: a high-scoring agent may produce unproductive descendants, while a lower-scoring one seeds lineages that achieve greater long-term gains. We term this phenomenon the \emph{Metaproductivity–Performance Mismatch}.

To address this mismatch, we introduce \emph{clade-level metaproductivity} ($\mathrm{CMP}$), inspired by Huxley’s notion of clades as lineages of common ancestry~\citep{huxley1957three}. $\mathrm{CMP}$ quantifies the productivity of a clade by aggregating the success of an agent’s descendants rather than relying solely on its immediate benchmark score. 
Furthermore, we show in Theorem~\ref{thm:main} that in our self-improving coding agent development setting (Assumption~\ref{assumption:main}, which includes the assumption that the only quality of the self-improvement process is the evaluation score of the final agent and that the evaluation is conducted with repeatable trials), having access to the true $\mathrm{CMP}$ oracle suffices to imitate the G\"odel Machine. 

This insight motivates our proposed algorithm, the \textbf{Huxley–G\"odel Machine (HGM)}, which approximates GM-style self-improvement by estimating $\mathrm{CMP}$ from clade-aggregated descendant outcomes and selecting nodes to expand via Thompson sampling. Furthermore, by leveraging a more reliable estimate, we adaptively decouple expansion from evaluation, leading to asynchronous execution for efficient parallelism.

Empirically, HGM better aligns with long-run agent productivity than benchmark-driven baselines, as shown in Figure~\ref{fig:intro} (left). 
On SWE-bench Verified~\citep{jimenez2024swebench,chowdhury2024swebenchverified} and Polyglot~\citep{Gauthier2024polyglot}, HGM consistently outperforms Darwin G\"odel Machine (DGM)~\citep{zhang2025dgm} and Self-Improving Coding Agent (SICA)~\citep{robeyns2025selfimprovingcodingagent}. 
Remarkably, one agent found by HGM surpasses SWE-agent~\citep{yang2024swe}, the highest-scoring human-engineered coding agent with officially checked results, on SWE-bench Lite~\citep{jimenez2024swebench}, when both use the GPT-5-mini backbone under matched budgets. 
The HGM-discovered agent transfers robustly when evaluated under a shift that is simultaneous in both the dataset and the model. Although optimized on SWE-bench Verified with GPT-5-mini, when tested on SWE-bench Lite with the GPT-5 backbone, \textbf{it achieves performance on par with the best officially verified human-engineered coding agents}.

To summarize, our contributions are as follows: 
\begin{itemize}[leftmargin=*,topsep=0em,itemsep=0em]
    \item We analytically define the Clade-Metaproductivity ($\mathrm{CMP}$) function as a measure of agents' self-improving ability and show that in a self-improving coding agent development setting (Assumption~\ref{assumption:main}), access to a $\mathrm{CMP}$ oracle suffices to reproduce the G\"odel Machine’s acceptance mechanism. (Theorem~\ref{thm:main}).
    \item We empirically observe that immediate benchmark performance is an unreliable predictor of $\mathrm{CMP}$ and show that our $\mathrm{CMP}$ estimator aligns better.
    \item Using our $\mathrm{CMP}$ estimator, we propose the \textbf{Huxley–G\"odel Machine (HGM)}, which approximates the G\"odel Machine in a coding agent setting from partial evaluations and guides the expansion via Thompson sampling with adaptive scheduling.
    \item We empirically validate HGM on SWE-bench Verified and Polyglot, demonstrating higher-quality optimized agents compared to previous self-improving methods, even though they were discovered within substantially smaller allocated CPU-hours. Furthermore, HGM \textbf{achieves human-level coding agent design} on SWE-bench Lite by optimizing on SWE-bench Verified. 
\end{itemize}

\section{Self-Improvement as Tree-Search}\label{sec:self-improvement}
Both the Darwin G\"odel Machine (DGM) and the Self-Improving Coding Agent (SICA) belong to the class of self-referential AI \citep{Schmidhuber1987EvolutionaryPI,Schmidhuber:03gm}. In particular, in DGM and SICA, agents modify themselves to generate new agents, each empirically validated on downstream tasks. 

In this paper, we formalize this self-improvement process as an iterative tree-search problem, where the goal is to discover an agent that maximizes performance across multiple downstream tasks. Concretely, starting from an initial agent as the root, a tree-search policy incrementally grows the tree of self-modified agents. At each iteration, the policy either selects an agent (a node in the tree) to expand by producing a child agent (a self-modified version of the selected agent) or selects an agent to undergo additional evaluation on downstream tasks.

Formally, let $\mathcal{T}_t$ denote the archive of our agents at iteration $t$. In this paper, the archive is always represented as a tree of evolved agents, and we use the terms archives and trees interchangeably.  
$\mathcal{T}_0 = \{a_0\}$ is initialized as a single-node tree with a fixed initial agent.
At each iteration $t$, the policy selects actions $a_{t+1} \sim \pi(\cdot \mid \mathcal{T}_t),$ where $\pi$ is a policy over actions $\mathcal{A}_t = \mathcal{M}_t \cup V_t$, $\mathcal{M}_t = \{m_a : a\in \mathcal{T}_t\}$ are agent modifications, and  $V_t =\{v_a : a\in \mathcal{T}_t\}$ denotes evaluations. 
Action $m_a$ instructs the agent $a$ to produce a self-modification that is added as a child of $a$ to the tree, and $v_a$ selects agent $a$ from the tree for an additional evaluation on one more downstream task.
After exhausting the computational budget, the policy selects a final agent ($a_{\text{final}} = \argmax_{a\in \mathcal{T}}Score_\pi(a) \in \mathcal{T}_B$ where $B$ is the termination iteration and $Score$ is part of the policy) from the final tree as the returned agent.
The objective is to optimize $J(\pi) = \E[U(a_{\text{final}})]$, where $U$ is a utility function that measures the performance of downstream tasks.
In this work, we consider $U$ the average of binary success indicators across all downstream tasks. $\pi$ denotes an algorithm, with DGM, SICA, and our proposed HGM representing concrete instances.

\paragraph{Compound Policy.} At each step of self-improvement, the system faces a compound decision: whether to expand the tree by generating new agents or to evaluate existing ones. This decision naturally decomposes into three sub-policies: (i) a \emph{selection policy} that chooses between expansion and evaluation, (ii) an \emph{expansion policy} that determines which parent to modify, and (iii) an \emph{evaluation policy} that selects an agent to test.  Prior approaches, such as SICA and DGM, conflate these choices. They always expand a parent, create a child, and immediately evaluate that child on multiple tasks. This fixed sequence restricts flexibility: once a new agent is generated, it monopolizes evaluations, even if older agents appear more promising. For instance, an agent that fails nine tasks in a row continues to consume evaluations, while an older agent with partial successes is ignored.

HGM breaks this rigidity by \emph{decoupling} expansion from evaluation. At each step, it adaptively decides whether to generate a new agent or to further probe an existing one, and evaluations are always at the granularity of a \emph{single agent–task pair}. This finer control enables early stopping on unpromising agents.
Table~\ref{tab:structured_policy} summarizes how SICA, DGM, and HGM instantiate these sub-policies.

\section{Huxley-G\"odel Machine}\label{sec:method}
In this section, we introduce the Huxley–G\"odel Machine (HGM), 
a self-improving machine that approximates G\"odel Machine by using clade-level statistics. At the core of HGM lies the notion of metaproductivity—a measure of an agent's ability to improve its self-improvement skills, which leads to better downstream performance of distant future agents.

In Section~\ref{sec:GM_SICAD} , we state the original G\"odel machine and describe how our scope is specialized with respect to the general setup of the G\"odel machine. In Section~\ref{sec:CMP}, we introduce two metrics of metaproductivity: Global metaproductivity ($\mathrm{GMP}$), which captures how evolving a given agent increases the metaproductivity of the entire agent tree. This measure of metaproductivity is general and difficult to operationalize or estimate. We instead introduce clade-metaproductivity ($\mathrm{CMP}$), which measures how promising the evolutions starting from a given agent (its clade) are. In Theorem~\ref{thm:main}, we show that access to true $\mathrm{CMP}$ is sufficient to implement a G\"odel Machine applied to the coding agent development setting (Assumption~\ref{assumption:main}).
Following on that, in Section~\ref {sec:main_alg}, we introduce the Huxley-G\"odel Machine (HGM), which guides the self-improvement search with Thompson Sampling based on the estimate of $\mathrm{CMP}$. 

\subsection{G\"odel Machine in the Self-Improving Coding Agent Development Setting}\label{sec:GM_SICAD}
The original G\"odel Machine is a general task solver that, in principle, can optimally make any provable self-improvements in any computable environment with respect to a given objective~\citep{Schmidhuber:03gm}. It achieves this by running a proof searcher, continually looking for formal proofs that some modification of its own code will yield higher expected utility. Once such a proof is found, the modification is executed and permanently alters the machine. Importantly, the theoretical analysis of the G\"odel Machine explicitly accounts for the fact that the agent has only a single life (with no repeatable trials) and that proving a self-improvement consumes real time and resources, which could be used for gathering the reward.

In contrast, this paper focuses on problems in a setting tailored to a particular kind of self-improving coding agent development that follows Assumption~\ref{assumption:main}. 
Specifically, our problem assumes that the only objective is the utility of the agent selected at the end of the agent development process.
Moreover, in our problem, the tests used to evaluate coding agents are conducted in a repeatable manner. This means that the testing environment is reset for each test, and prior evaluations do not influence later test results. Furthermore, each self-modification reduces the remaining budget by exactly one unit. For the theoretical analysis, we also assume that the only operation that incurs a time cost is the self-modification itself.

Within this framework designed for self-improving coding agent development, the G\"odel Machine is an optimal agent operating in a POMDP where the policy observes only a parent $a_\text{parent}$, its child $a_\text{child}$, and the remaining budget $b$, and then chooses to accept or reject the child. At termination, $Score_\pi$  selects either the final parent or child as output. Full description of the POMDP can be found in the Appendix~\ref{app:GM_CMP}. We clarify the additional assumptions in our setup in comparison to the general G\"odel Machine in Assumption~\ref{assumption:main}.

\begin{assumption}\label{assumption:main}
For the theoretical analysis of G\"odel Machine applied to self-improving coding agents, we make the following additional assumptions in comparison to the setup from the original G\"odel Machine:
\begin{itemize}
    \item The policy objective function is defined as a function of only the final agent, with no other rewards received before termination;
    \item The agent’s utility is measured by its performance on evaluation tasks, under the assumption of repeatable trials: for any agent–task pair, the expected outcome is independent of evaluation time or prior events.
    \item The proofs of G\"odel Machines do not consume budget;
    \item And each self-modification costs exactly one unit of the budget.
\end{itemize}
\end{assumption}

\subsection{Metaproductivity and Clade-Metaproductivity}\label{sec:CMP}

The global metaproductivity measures the impact of self-improvement on the entire current tree archive of all agents. 
Given a policy $\pi$, to quantify the quality of how an agent's self-modification influences the performance of the system, we define the notion of global metaproductivity ($\mathrm{GMP}$):
\[
\mathrm{GMP}_\pi(\mathcal{T},a) \;=\; 
\mathbb{E}_{\mathcal{T}_B \sim p_\pi(\cdot \mid \mathcal{T},a)}
\left[   U(\text{argmax}_{a' \in \mathcal{T}_B}Score_\pi(a')) \right],
\]
where $\mathcal{T}$ is a tree of agents and $a\in\mathcal{T}$. $Score_\pi$ is the function that scores the agents for the final selection. The policy $\pi$ unrolls the trajectory until the end of the episode with policy $\pi$ and produces a final archive of agents $\mathcal{T}_B$. The distribution of the trajectory is given by $p_\pi$.

$\mathrm{GMP}$ directly corresponds to the Q-value function in reinforcement learning, with the state phrased as the archive of agents and the action being the selected agent to expand. The $\mathrm{GMP}$ value of a node measures how well (in expectation) the final agent obtained from the search process will perform. $\mathrm{GMP}$ measures the long-term potential of self-improvement for the entire tree, which also includes modifications that improve self-improvement itself, and so on. 
An algorithm might, at the beginning, focus on improving the ability to self-improve while neglecting direct benchmark abilities, only to later focus on them. 
This is a principal meta-learning behavior that is captured in the original G\"odel Machine~\citep{Schmidhuber:03gm}. The objective of designing a policy for self-improvement (Section~\ref{sec:self-improvement}) is equivalent to optimizing $\mathrm{GMP}(\{a_0\}, a_0)$.

While $\mathrm{GMP}$ captures the full global potential of a policy with access to the history of agents, its scope is too broad for practical conceptualization, since, in principle, the self-modifications of an agent can influence the expected utility of a parent agent by introducing new information.
The G\"odel Machine achieves global-optimality by deciding whether to accept or reject self-modification, focusing on the provable potential subsequent self-improvements.
Motivated by this observation, we define a localized variant of $\mathrm{GMP}$ that focuses on the subtree rooted at a given agent, i.e., its \emph{clade}. 
We refer to this quantity as Clade-Metaproductivity ($\mathrm{CMP}$):
\begin{align*}
\mathrm{CMP}_\pi(\mathcal{T},a) &= \mathbb{E}_{\mathcal{T}_B \sim p_\pi(\cdot \mid \mathcal{T},a)}
\left[  U(\text{argmax}_{a' \in C(\mathcal{T}_B, a)}Score_\pi(a')) \right]\\
&=\mathbb{E}_{\mathcal{T}_B \sim p_\pi(\cdot \mid \mathcal{T},a)}
\left[  \text{max}_{a' \in C(\mathcal{T}_B, a)}U(a') \right] \qquad \text{(if $Score = U$)},\\
\end{align*}    
 where $C(\mathcal{T}_B, a)$ is the clade (i.e., the subtree with $a$ as the root) of the node $a$ in the Tree $\mathcal{T}_B$ and $Score$ is the final agent selection metric.
 $\mathrm{CMP}$ contains the non-greedy information about the future evolution of self-improving agents, therefore guiding good strategies for self-improvement aimed also at the improvement of self-improvement itself. Furthermore, we show the crucial relation of $\mathrm{CMP}$ to the G\"odel Machine.

\begin{theorem}\label{thm:main}
Under Assumption~\ref{assumption:main}, access to the $\mathrm{CMP}$ oracle is sufficient to implement the G\"odel Machine.
\end{theorem}

The proof is available in the App.~\ref{app:GM_CMP}. This observation motivates us to introduce the estimate of $\mathrm{CMP}$ and use this as guidance in our algorithm. 
By estimating $\mathrm{CMP}$, HGM approximates the G\"odel Machine. We describe our algorithm fully in the next section.

\subsection{Algorithm}\label{sec:main_alg}

 Existing methods use benchmark performance on coding tasks as a guiding metric, treating task success as an indicator of self-improvement potential. This assumption is overly greedy: it evaluates only the immediate utility of a modification while ignoring its downstream consequences for future self-modifications. We refer to this gap as the \emph{Metaproductivity-Performance Mismatch}: the divergence between short-term task performance and the long-term capacity for self-improvement, as measured by $\mathrm{CMP}$. Empirical evidence shows that this mismatch occurs in practice (see Section~\ref{sec:metaproductivity-performance-mismatch}.) 
We aim to model long-term, global dependencies by deriving our estimator of $\mathrm{CMP}$. Specifically, we define HGM by stating its three subpolicies.

\paragraph{Expansion Policy.}\label{method:expand-policy}

 The core of the HGM algorithm is its selection criterion for expansion. HGM aims to estimate Clade-Metaproductivity with the motivation that, in our setting, under Assumption~\ref{assumption:main}, the true $\mathrm{CMP}$ as the criterion would produce the G\"odel Machine due to Theorem~\ref{thm:main}. In this sense, HGM approximates G\"odel Machine, the \textbf{optimal} self-improving machine. This is in contrast to the currently used greedy selection criteria based on performance metrics, which ignore the potential of the model to improve its self-improving abilities.

We estimate $\mathrm{CMP}$ using the \textbf{weighted} average of agents' empirical performance in the clade. 
HGM is designed to promote higher weights to higher utility agents. See below for how our evaluation policy induces this weighting strategy.
Formally, let us assume a fixed archive of agents $\mathcal{T}_t$, $n_{\text{success}}(a)$ be the number of passed tests of $a$, and $n_{\text{failure}}(a)$ be the number of failed tests of $a$. Then
\[
n^C_\text{success}(a) = \sum_{a' \in C(a)} n_\text{success}(a') \quad \text{and} \quad n^C_\text{failure}(a) = \sum_{a' \in C(a)} n_\text{failure}(a'),
\]
where $C(a)$ is the clade of $a$ in $\mathcal{T}_t$. We define our Clade-Metaproductivity estimator as
\[{\widehat{\mathrm{CMP}}(a)} =\frac{n^C_\text{success}(a)}{n^C_\text{success}(a)+n^C_\text{failure}(a)}.\]

Evaluating productivity at the level of entire clades rather than individual agents offers several key advantages. It aligns better with the goal of self-improvement, as a modest ancestor can still be highly valuable if its descendants consistently advance, while stagnant lineages are deprioritized. At the same time, aggregating evidence across a clade yields more statistically robust estimates than single-node outcomes by using information from more samples. This is particularly important when evaluations are costly and benchmarks are only partially observed. 

$\widehat{\mathrm{CMP}}(a)$ can be viewed as a weighted sum over the empirical means of agents in $C(a)$, with the weight for an agent being the number of task evaluations it has performed. Furthermore, we design our evaluation selection in such a way that it selects \textbf{highly performing agents}, which creates a selection of a soft maximum in the clade. 

After calculating the $\mathrm{CMP}$ estimates, the HGM probabilistically approximates the selection of the highest scoring agent using Thompson Sampling—a standard method in the bandit literature for smoothly maximizing the decision criterion~\citep{agrawal2012analysis,chapelle2011empirical}. We will refer to $a \sim TS(\{(n_s, n_f) | n\in \mathcal{T}_t \})$ as the agent sampled from the Thompson-Sampling process with parameters $n_s$ (number of successes) and $n_f$ (number of failures).
Given that the search problem has a known budget, our algorithm introduces an exploration-exploitation scheduler $\tau$ that is monotonically increasing with respect to the current time $t$, encouraging exploration in the early stages and polarization of the sampling distribution as it approaches the end. Formally, we select the agent to expand $a^*$ as 
\[
a^* \sim TS(\{(\tau(1+n_\text{success}^C(a)), \tau(1+n_\text{failure}^C(a))) | a\in \mathcal{T}_t \}).
\]

\paragraph{Evaluation Policy.}\label{method:eval-policy}
As stated in the expansion policy, we design our evaluation policy to prioritize \textbf{agents} with a higher evaluation score to induce the selection of the maximum over the clade. Formally, the agent to evaluate $a^*$ is sampled from the Thompson Sampling process with 
\[
a^* \sim TS(\{(\tau(1+n_\text{success}(a)), \tau(1+n_\text{failure}(a)) ) | a\in \mathcal{T}_t\}.
\]

\paragraph{Selection Policy.}

Finally, our agent has to choose between expansion and evaluation.
At each iteration, the algorithm first selects whether to evaluate or expand. Previous methods have evaluated newly created agents directly after their creation. 
Our novel estimation of agent self-improving quality has the additional benefit of collecting more samples more quickly (because it has samples from the entire clade). 
This enables more fine-grained control over when to evaluate and when to create a new agent for better efficacy. 
Therefore, we decouple evaluation from expansion and treat them as separate steps.

To decide how and when to evaluate or expand agents, we draw inspiration from the infinite-armed bandit literature. Infinite-armed bandit problems capture the tension between repeatedly sampling known options to reduce uncertainty about promising arms and exploring new options that have the potential to perform better. This perspective provides a natural lens for our setting, where evaluations correspond to sampling existing arms, and expansions correspond to introducing new ones.
In this work, we follow the strategy of UCB-Air~\citep{NIPS200849ae49a2}, which adds arms when the number of evaluations $N^\alpha \geq m$ for some $\alpha \in [0, 1]$, where $m$ is the number of existing arms. In our case, arms correspond to the agents; hence, we decide to expand at time $t$ if $N_t^\alpha \geq |\mathcal{T}_t|$.

\paragraph{Final Agent Selection Strategy.}\label{method:final-agent-select}
HGM iteratively executes the structured policy defined by our selection policy, expansion policy, and evaluation policy. When the computational budget exceeds, it returns the agent with the highest $\epsilon$ percentile of the utility posterior in the final tree for some hyperparameter $\epsilon$, namely the \textbf{best-belief agent}. Formally, a best-belief agent is defined as
\[
\text{argmax}_{a\in \mathcal{T}_B} I_\epsilon(1 + n_{\text{success}(a)}, 1 + n_{\text{failure}}(a)),
\]
where $I$ is the regularized incomplete beta function. See Algorithm\ref{alg:hgm} in Appendix~\ref{app:alg} for the detailed procedure of HGM.

\paragraph{Asynchronous Implementation.}\label{method:async-parallel}
As an additional benefit of decoupling the policy, we introduce asynchronous execution of evaluation and expansion. Since the execution of coding agents generally requires querying large language models multiple times, the computation time can be lengthy. To boost our algorithm, we propose the asynchronous HGM algorithm (HGM Async), which utilizes all available computational power until the computational budget is exceeded. HGM Async concurrently executes an iteration of the process on each available CPU. Once one iteration finishes, a new iteration immediately starts. It uses the most recent data, with one exception, and updates the data once it is finished. The exception is that one needs to take all running expansions and explorations into consideration when executing the selection strategy. 
See experimental results~\ref{tab:self-improv-compare} for the runtime comparison with DGM and SICA.

\section{Experimental Results}
\label{sec:exp}
We evaluate HGM on challenging software engineering tasks to assess three core aspects: \textbf{1)} the fidelity of HGM’s $\mathrm{CMP}$ estimation (Sec.~\ref{sec:metaproductivity-performance-mismatch}), \textbf{2)} its capability for self-improvement with HGM compared with DGM and SICA (Sec.~\ref{exp:eval-self-improv}), and \textbf{3)} the effectiveness in automatic agent design through evolutionary processes, benchmarked against a leading human design up to date\footnote{The leading SWE-agents on \url{https://www.swebench.com} (Lite) as of 24 October 2025.} (Sec.~\ref{exp:vs-human}). We conducted our experiments on the SWE-bench Verified (SWE-Verified) and SWE-bench Lite (SWE-Lite) variants, as well as the Polyglot problems, both of which consist of coding challenges and are widely used for coding agent evaluation~\citep{xia2025demystifying,zhang2024autocoderover,zhang2025swe}. We follow DGM's evaluation setting for Polyglot problems, where agents have no access to private test cases or test results.  For budget considerations, in addition to the full datasets, we use 60-task subsets (SWE-Verified-60), derived from the first two stages of DGM’s progressive evaluation. In all experiments, we employ HGM with an exploration-exploitation scheduler $\frac{B}{b}$, where $b$ is the remaining budget, $\epsilon = 1$, and $\alpha = 0.6$. All experiments involving HGM use the HGM‑Async algorithm.
We apply an identical initial agent when compared to DGM and SICA, which is adopted from the official implementation of DGM. See Appendix~\ref{app:exp} for a detailed description of the initial agents used in different experiments.

\subsection{Metaproductivity-Performance Mismatch}
\label{sec:metaproductivity-performance-mismatch}
The experiments in this section are designed to serve two purposes: (i) to provide evidence of the Metaproductivity-Performance Misalignment (MPM) issue; and (ii) to assess whether the $\widehat{\mathrm{CMP}}$ of HGM is a more reliable $\mathrm{CMP}$ estimator than the utility measures adopted by DGM and SICA. 
To reveal the misalignment, we compute the correlation between their selection criterion and empirical $\mathrm{CMP}$. 
To obtain empirical $\mathrm{CMP}s$, we analyze the expanded search tree after each method has completed its run. For every node in the tree, we define its empirical $\mathrm{CMP}$ as the maximum empirical mean of the task performance achieved within its clade, with the root of this clade excluded. 
This construction ensures that empirical $\mathrm{CMP}$ captures the self-improvement ability of a node.
We exclude the root of a clade to avoid the circular use of the target in the estimators.
For HGM, the $\widehat{\mathrm{CMP}}$ is defined as a function over the clade of a node; a critical adjustment is required to avoid target leakage. 
Specifically, we exclude the evaluations that are most directly related to the target: the root of the clade (an ancestor of the target) and the subtree rooted at a direct child of the evaluated node that contains the empirical maximum, thereby ensuring a fair comparison (see Appendix~\ref{app:cmp_estimation} for detailed computation). 
We report both the correlation coefficient weighted by the number of evaluations used in prediction, as well as the unweighted correlation.
We conducted our experiments on the SWE-Verified-60 and Polyglot datasets.

\begin{table*}[t!]
  \centering
  \captionof{table}{
  \textbf{Clade-Metaproductivity: Empirical vs. Estimation Correlation}. We report the Pearson correlations between the empirical $\mathrm{CMP}s$ and the estimates from DGM, SICA, and HGM on SWE-Verified-60 and Polyglot. For the weighted correlations, each prediction is weighted by the number of evaluations it has accessed.
  }\label{tab:vs-corr}
  \renewcommand\arraystretch{1.2} 
  \resizebox{0.9\linewidth}{!}{
    \begin{tabularx}{\linewidth}{
      l|
          >{\columncolor{lightwheat!70}}Y
          >{\columncolor{skyblue!20}}Y|
          >{\columncolor{lightwheat!70}}Y
          >{\columncolor{skyblue!20}}Y
    } 
      \toprule
      \multirow{2}{*}{\textbf{Estimates}} 
        & \multicolumn{2}{c|}{\textbf{SWE-Verified-60}} 
        & \multicolumn{2}{c}{\textbf{Polyglot}} \\
      \cmidrule(lr){2-3}\cmidrule(l){4-5}
        & \multicolumn{1}{c}{\textbf{Weighted}} & \multicolumn{1}{c}{\textbf{Un-weighted}} 
        & \multicolumn{1}{c}{\textbf{Weighted}} & \multicolumn{1}{c}{\textbf{Un-weighted}} \\ 
      \midrule
      SICA & 0.444    & 0.444      & 0.274     & 0.274 \\
      DGM  & 0.285    & 0.406      & 0.383     & 0.357 \\ 
      \hline
      \textbf{HGM (Ours)} & \textbf{0.778} & \textbf{0.512}  & \textbf{0.626} & \textbf{0.873}  \\
      \hline
      \bottomrule
    \end{tabularx}
  }
\end{table*}

\textbf{Results \& Discussion.} Table~\ref{tab:vs-corr} summarizes the correlations between the three estimators and the empirical $\mathrm{CMP}s$ (the targets). We first observe that the SICA and DGM estimators achieve positive Pearson correlation coefficients: 0.444 and 0.285 on SWE-Verified-60, and 0.274 and 0.383 on Polyglot, respectively, suggesting weak alignments, i.e., MPM. 
In comparison, HGM's estimator, $\widehat{\mathrm{CMP}}$, achieves substantially stronger weighted correlations of 0.778 and 0.626 on SWE-Verified-60 and Polyglot, respectively, as well as unweighted correlations of 0.512 and 0.8783, surpassing SICA and DGM by wide margins.
These results provide strong indications that HGM, equipped with $\widehat{\mathrm{CMP}}$, offers a more reliable estimate of metaproductivity and effectively mitigates the MPM issues inherent to SICA and DGM.

\subsection{Evaluating HGM Self-Improving Capability}
\label{exp:eval-self-improv}
To validate our hypothesis that our $\mathrm{CMP}$ estimator better predicts future self-improvement and, hence, leads to more effective self-modifications, we evaluate HGM against two state-of-the-art self-improving coding agent methods: DGM and SICA. We conduct controlled experiments under the same setup as DGM, with a budget-friendly modification applied consistently to both HGM and all baselines to ensure fairness, i.e., we adopt more cost-efficient backbone LLMs (GPT-5 for expansion and GPT-5-mini for evaluation for SWE-Verified; Qwen3-Coder-480B-A35B-Instruct for expansion and Qwen3-Coder-30B-A3B-Instruct for evaluation for Polyglot). For all methods, we start with the same initial ancestor, which achieves 40\% and 20.3\% on the SWE-Verified-60 and Polyglot, respectively. 
We compare the task performance of their best-belief final agents after a maximum allowance of 800 benchmark task evaluations, selected in DGM and SICA using empirical means.
In addition, our asynchronous parallelization of expansion and evaluation enables self-improvement to consume fewer allocated CPU-hours than DGM and SICA (see Sec.~\ref{method:async-parallel}). We verify this and report the allocated CPU-hours required for 800 evaluations.

\begin{table*}[t!]
  \centering
  \captionof{table}{
  \textbf{Self-Improving Capability Comparison}. We report the task performance (in accuracy) of each method’s best-belief agent and the allocated CPU-hours time required for 800 evaluations. Superscripted accuracies with ``$+$'' indicate performance gains over their respective initial agents. 
  }
  \label{tab:self-improv-compare}
  \renewcommand\arraystretch{1.2} 
  \resizebox{0.9\linewidth}{!}{
    \begin{tabularx}{\linewidth}{
      l|
          >{\columncolor{lightwheat!70}}Y
          >{\columncolor{skyblue!20}}Y|
          >{\columncolor{lightwheat!70}}Y
          >{\columncolor{skyblue!20}}Y
    } 
      \toprule
      \multirow{2}{*}{\textbf{Best-belief Agent of}} 
        & \multicolumn{2}{c|}{\textbf{SWE-Verified-60}} 
        & \multicolumn{2}{c}{\textbf{Polyglot}} \\
      \cmidrule(lr){2-3}\cmidrule(l){4-5}
        & \multicolumn{1}{c}{\textbf{Acc. (\%)$\uparrow$}} & \multicolumn{1}{c}{\textbf{Time (hours)$\downarrow$}} 
        & \multicolumn{1}{c}{\textbf{Acc. (\%)$\uparrow$}} & \multicolumn{1}{c}{\textbf{Time (hours)$\downarrow$}} \\
      \midrule
      SICA & 50.0$^{+10}$     & infinite loop       & 25.4$^{+5.1}$     & 572 \\
      DGM  & 53.3$^{+13.3}$   & 1231      & 27.1$^{+6.8}$     & 2385 \\ 
      \hline
      \textbf{HGM (Ours)} & \textbf{56.7}$^{+16.7}$ & \textbf{517} & \textbf{30.5}$^{+10.2}$ & \textbf{347} \\
      \hline
      \bottomrule
    \end{tabularx}
  }
\end{table*}

\textbf{Results \& Discussion.} We summarize the comparison results in Table~\ref{tab:self-improv-compare}. Across both SWE-Verified-60 and Polyglot, all three methods successfully perform agent discovery by optimizing the initial agent through self-improvement. However, HGM’s best-belief agent demonstrates not only the highest task performance---56.7\% on SWE-Verified-60 and 30.5\% on Polyglot---but also the best efficiency, requiring the fewest allocated CPU-hours for 800 evaluations: 6.86 times faster than DGM and 1.65 times faster than SICA on Polyglot, and 2.38 times faster than DGM on SWE-Verified-60. Notably, on SWE-Verified-60, SICA repeatedly encounters ``query length out-of-LLM-context-window'' during self-improvement processes after 360 evaluations. Despite this, the Polyglot results validate our hypothesis regarding HGM’s runtime advantage over the baselines. In conclusion, HGM, equipped with a better utility estimator and asynchronous expansion–evaluation iterations, establishes itself as a more effective self-improving mechanism compared to DGM and SICA.

\subsection{HGM vs. Humans: on Coding Agents Design}
\label{exp:vs-human}
To gain a better understanding of its potential, we extend our evaluation of HGM by benchmarking it against the best human performance in coding agent design on SWE-Lite. We consider two settings: 1) optimization on full SWE-Verified and 2) generalization to SWE-Lite.

\subsubsection{Optimization on Full SWE-Bench Verified} \label{exp:vs-human-verified}
In this experiment, rather than using the SWE-Verified-60, we scaled the HGM evaluation to the full SWE-Bench Verified benchmark (500 coding challenges) with an increased number of HGM iterations (8000 evaluations). 
We further adjusted the initial agent so that it yields an improved accuracy of $53.2\%$. See Appendix~\ref{app:exp} for the adjustment details.
Notably, this stronger starting point underscores the difficulty of further improvement due to a higher baseline. Despite this, HGM demonstrates significant gains and strong absolute performance.

\textbf{Results \& Discussion.} After 8000 evaluations, HGM discovered an optimized agent that solves $61.4\%$ tasks, surpassing the best human-designed agent built on GPT-5-mini on the SWE-Verified leaderboard. This establishes our discovered agent as the \textit{top-scoring} GPT-5-mini–based system, and positions it among the \textit{top-10} agents over all checked submissions, even when compared to systems built on stronger backbone models that can cost $5\times$ more (e.g., Claude-3.7). While higher scores on the leaderboard do not necessarily indicate superior general coding ability---since both human- and machine-designed agents may overfit to the benchmark---these results demonstrate a promising potential of HGM for competing with established human-designed baselines under identical model constraints. 

\subsubsection{Generalization To SWE-Bench Lite} \label{exp:vs-human-lite}
To validate that HGM’s self-evolution produces agents with stronger general coding ability, rather than merely overfitting to SWE-Verified, we evaluate the top agent discovered on SWE-Verified against unseen tasks. Specifically, we compare this agent with its initial ancestor (which achieved $53.2\%$ on SWE-Verified) using SWE-Lite, a benchmark of 300 coding tasks, 93 of which overlap with SWE-Verified. For rigor and comparability, we report two settings: (i) a filtered setting where the 93 overlapping tasks are excluded, leaving only completely unseen tasks, and (ii) the full 300-task benchmark, identical to the standard evaluation used for human designs on the leaderboard. As of the time of writing, no checked submission using GPT-5-mini appears on the SWE-Lite leaderboard. To control for backbone differences and isolate agent design, we adapt the leading system (with checked submissions) (SWE-agent + Claude 4 Sonnet) by replacing its backbone with GPT-5-mini, yielding SWE-agent + GPT-5-mini, as an additional baseline for comparison.

\begin{table*}[t!]
  \centering
  \captionof{table}{
  \textbf{Generalization on SWE-Lite: HGM's Best-belief SWE-Verified Agent.} We report the accuracy of HGM’s best-belief SWE-Verified agent on SWE-Lite under two settings: filtered (excluding tasks overlapping with SWE-Verified) and standard (the official leaderboard setting used for evaluating human-designed agents)).
  }\label{tab:vs-human-lite}
  \renewcommand\arraystretch{1.2}
  \resizebox{0.9\linewidth}{!}{
  \begin{tabularx}{\linewidth}{
      l|
          >{\columncolor{lightwheat!70}}Y|
          >{\columncolor{skyblue!20}}Y
    } 
    \toprule
    \multicolumn{1}{c|}{\textbf{Coding Agents}}   & \multicolumn{1}{c|}{\textbf{SWE-Lite Filtered (\%)}} & \multicolumn{1}{c}{\textbf{SWE-Lite Standard (\%)}} \\
    \hline
      HGM Initial Ancestor    & 34.8   & 44.0 \\
      SWE-agent+GPT-5-mini    & 39.6   & 47.6 \\
    \hline
     \textbf{HGM's Best-belief SWE-Verified Agent} & \textbf{40.1}   & \textbf{49.0} \\ %\silvermedal} \\ 
    \hline    
    \bottomrule
  \end{tabularx}
}
\end{table*}

\renewcommand{\arraystretch}{1.8}
\begin{table*}[t!]
  \centering
  \captionof{table}{
  \textbf{Transfer to different LLMs on SWE-Lite: HGM's Best-belief SWE-Verified Agent.} Similarly, We report the accuracy of HGM’s best-belief SWE-Verified (optimized with GPT-5-mini) agent on SWE-Lite (evaluated with GPT-5) under two settings: filtered (excluding tasks overlapping with SWE-Verified) and standard (the official leaderboard setting used for evaluating human-designed agents).
  }\label{tab:gpt5-vs-human-lite}
  \renewcommand\arraystretch{2}
  \resizebox{0.9\linewidth}{!}{
  \begin{tabularx}{\linewidth}{
      l|
          >{\columncolor{lightwheat!70}}Y|
          >{\columncolor{skyblue!20}}Y
    } 
    \toprule
    \multicolumn{1}{c|}{\textbf{Coding Agents}}   & \multicolumn{1}{c|}{\textbf{SWE-Lite Filtered (\%)}} & \multicolumn{1}{c}{\textbf{SWE-Lite Standard (\%)}} \\
    \hline
    SWE-agent (Best on the LB)    & \textbf{48.3}   & 56.7 \\
    \hline
     \parbox[c][5ex][c]{5.9cm}{
      \textbf{HGM's Best-belief SWE-Verified Agent + GPT-5} 
     }
     & 47.8   & \textbf{57} \\ %\silvermedal} \\ 
     \hline
    \bottomrule
  \end{tabularx}
}
\end{table*}

\textbf{Results \& Discussion.} We show the generalization results of HGM's best-belief SWE-Verified agent on SWE-Lite benchmark in Table~\ref{tab:vs-human-lite}. The best-belief HGM agent found on SWE-Verified achieves $40.1\%$ under the filtered (completely unseen) setting and $49.0\%$ under the standard setting. Compared to its initial ancestor ($34.8\%$ and $44.0\%$, respectively), these gains substantiate the effectiveness of HGM’s self-evolution in improving general coding ability---rather than overfitting to the optimization set. Notably, the superior performance of our HGM agent achieved on the standard SWE-Lite places it firmly \textit{in second place} on the SWE-Lite leaderboard among all checked submissions. Moreover, based on our local execution result of SWE-agent using the SWE-agent + Claude 4 Sonnet submission version with the same configuration, the agent optimized by HGM outperforms SWE-agent + GPT-5-mini, which achieves $39.6\%$ (vs. $40.1\%$ for us) on the filtered and $47.6\%$ (vs. $49.0\%$ for us) on the standard. This demonstrates that the edge arises not from the GPT-5-mini backbone, but from the genuine design improvements introduced by HGM evolution.

\textbf{Transfer to bigger LLMs.} We also examined how the discovered agent scales when paired with larger and better-performing LLMs. In particular, we replaced the GPT-5-mini backbone of HGM's Best-belief agent with the GPT-5 model to test whether the agent optimized with one LLM remains effective with another. The results indicate that the agent maintains its strong performance under this transfer. In particular, its accuracy on the SWE-Lite benchmarks is comparable to that of state-of-the-art, human-engineered coding agents reported on the leaderboard, suggesting that HGM’s self-evolved design principles are reliably transferred across backbone sizes. The agent discovered by HGM with GPT outperforms all other agents on the SWE-Bench Lite leaderboard with officially checked results and is one task behind the best-performing model on our selected "SWE-bench Filtered". In Table~\ref{tab:gpt5-vs-human-lite}, we compare it with the SWE-agent, which holds first place on the official SWE-Bench Lite leaderboard at the time of publication. This result further supports the conclusion that the improvements are due to genuine agent improvement rather than overfitting to a particular dataset or LLM.

\section{Related Works}

The general concepts of machine self-improvement were first systematically articulated by \citet{good1966speculations}, who described the possibility of ``Intelligence Explosion" once machines acquire the capacity to design more capable successors. Early work on explicit self-improvements dates back to \citet{Schmidhuber1987EvolutionaryPI}, which introduced self-referential learning mechanisms in which a system generates and evaluates modified descendant versions of itself.
Follow-up work on self-improvement progressed through interaction and agentic reinforcement learning. 
The Success-Story Algorithm(SSA)~\citep{schmidhuber1996multi, schmidhuber1997shifting}  progressively forces self-modifying policies to discover more effective self-modification strategies. Its core mechanism is based on hindsight: at each checkpoint, a sequence of self-modifications that did not yield higher long-term reward rates is systematically undone. In this way, SSA enforces continual improvement by ensuring that only those self-modifications associated with demonstrably greater reward intake per unit time are preserved.
Fitness-Monotonic Execution~\citep{kirsch2022self,kirsch2022eliminating} reduces the outer-loop design by favoring the execution of models with higher ancestral performance. Meta-discovered update rules optimized optimizers~\citep{metz2021training} and black-box search~\citep{lange2023discovering}. 
On the other hand, the G\"odel Machine, a fully self-referential algorithm that rewrites its own code whenever it can \emph{prove} an expected-utility improvement, provides a provably and globally optimal mechanism for self-improvement~\citep{Schmidhuber:03gm}.

The rise of contemporary LLMs has created an opportunity to automate substantial aspects of software engineering. One concrete step in this direction is the development of coding agents, which extend LLMs with the ability to operate in conventional computing environments. ChatDev~\citep{qian2023chatdev} first illustrated this idea in the context of automated bug fixing, and similar frameworks were later explored in SWE~\citet{yang2024swe}, OpenHands~\citep{wang2024openhands}, MetaGPT~\citep{hong2024metagpt}, and AgentLess~\citep{xia2025demystifying}.

The Self-Taught Optimizer~\citep{zelikmanSelfTaughtOptimizerSTOP2024} and G\"odel Agent~\citep{yin2024g} first experimented with agents that modify their own scaffolding.
Subsequently, DGM~\citep{zhang2025dgm} and SICA~\citep{robeyns2025selfimprovingcodingagent} extend this direction by implementing self‑modifying machines as full software engineering projects, where agents self-reference and modify their own repositories while validating changes through execution‑grounded software engineering tasks.
Both DGM and SICA, explicitly or implicitly, assume that higher software benchmark scores correspond to greater self‑improvement capacity. In contrast, HGM introduces a qualitative measure of self‑improvement consistent with the theoretical G\"odel Machine and directs self‑modifications using estimates of this measure.

The identified tree-search problem spans fixed-budget best-arm identification (BAI), Monte Carlo Tree Search, and infinite-armed bandits, introducing a distinct decision: explicit expansion actions that create new candidate leaves alongside ordinary evaluations. Fixed-budget BAI and Bayesian value-of-information methods assume a finite and known set of arms and offer guarantees for static candidates, thus not modeling the discovery of unknown arms \citep{audibert2010best,karnin2013almost,frazier2008knowledge}. Monte-Carlo Tree Search and its UCT variants~\citep{coulom2006efficient,kocsis2006bandit} alternate selection, expansion, and simulation, while their backup and selection rules typically target cumulative reward rather than fixed-budget final-choice objectives under noisy, low-signal feedback, with limited guarantees for pure exploration of leaf quality \citep{kaufmann2017monte}. Infinite-armed bandit formulations capture the explore-discover tradeoff but typically model discoveries as i.i.d. draws from a reservoir, missing tree structure, and hierarchical dependencies \citep{NIPS200849ae49a2,bubeck2011pure,carpentier2015simple}.

\section{Conclusion}
In this work, we identify a key limitation in the search heuristics of current self-improving coding agents: Benchmark scores alone do not reliably indicate an agent’s long-term potential for self-improvement, since high-scoring agents can still lead to stagnating lineages, while seemingly weaker ones may seed productive self-improvements. We refer to it as the Metaproductivity–Performance Mismatch. To address this gap, we introduce Clade-Metaproductivity ($\mathrm{CMP}$), a lineage-based metric inspired by Huxley’s notion of clades.
We show that, under certain assumptions, when applied to our self-improving coding agent search problem (Assumption~\ref{assumption:main}), the $\mathrm{CMP}$ oracle is sufficient to implement the G\"odel Machine (Theorem~\ref{thm:main}).

Building on this principle, we propose the Huxley–G\"odel Machine (HGM), which approximates $\mathrm{CMP}$ and uses it to guide expansion through Thompson sampling with adaptive scheduling. Empirically, HGM consistently produces higher quality agents than prior self-improving frameworks while also reducing wall-clock time. Notably, HGM generalizes across both dataset and model shifts, achieving human-level coding agent design performance on SWE-bench Lite with GPT-5 despite being optimized on SWE-bench Verified with GPT-5-mini.

Taken together, these results suggest that clade-based measures of improvement potential, rather than immediate performance alone, lead to more effective forms of self-improvement. By demonstrating that clade-level evaluation can reliably guide the growth of coding agents, this work points to a new paradigm for the design of agentic improvement: one in which improvement is driven not by narrow benchmarks, but by the long-term generative potential of entire lineages. This perspective underscores the importance of systems that strengthen an agent’s capacity to keep improving over time, rather than merely boosting their performance in the short term.

\section*{Acknowledgment}
We thank Yuhui Wang for the discussions during the early stages of this project. We gratefully acknowledge Jenny Zhang and Shengran Hu, the authors of Darwin G\"odel Machine, for sharing their insights about DGM and their implementation experience. We also thank Yilan Zhang, Rui Zhang, and Lisiyu Xie for their help in designing the visualizations.
The research reported in this publication was supported by funding from King Abdullah University of Science and Technology (KAUST) - Center of Excellence for Generative AI, under award number 5940.

\bibliography{arxiv}
\bibliographystyle{arxiv}

\appendix

\newpage

\section{G\"odel Machine with $\mathrm{CMP}$ Oracle}~\label{app:GM_CMP}

\textit{Assumptions. }The original G\"odel Machine is defined in a time-aware setting, where the prover must establish not only that a proposed self-modification increases expected objectives, but also that this improvement still holds after accounting for the time required to search for proofs and compute the modification. This is necessary because, in the general case, the environment may change during these computations, and the objective is measured with respect to elapsed time.

In our setup, by contrast, the environment and the evaluation metric remain fixed throughout the agent’s execution. The benchmark does not evolve over time, and the utility of any given agent is determined solely by its final performance on this static task. Importantly, we assume that the utility is measured by evaluation on tasks. It also follows the assumption of repeatable trials, meaning that the evaluation of a given agent on a task is independent of evaluation time or prior events. In other words, we are able to reset the testing environment for each test.
Furthermore, we assume that the G\"odel Machine prover has full knowledge of the utility function as part of its axioms of the environment. Hence, we exclude the evaluation actions from the action space. 
Finally, we assume that the G\"odel Machine prover does not consume budget, and that the self-modifications consume an equal amount of budget—exactly one budget unit.

To summarize, we show that the $\mathrm{CMP}$ oracle is sufficient to imitate G\"odel Machines in our specific setting that satisfies the following: \begin{itemize}
    \item The policy objective function is defined as a function of only the final agent, with no other rewards received before termination;
    \item The agent’s utility is measured by its performance on evaluation tasks, under the assumption of repeatable trials: for any agent–task pair, the expected outcome is independent of evaluation time or prior events.
    \item The G\"odel Machine prover has full knowledge of the utility function;
    \item The computation budget is finite and known;
    \item The proofs of G\"odel Machines do not consume the budget;
    \item And each self-modification costs exactly one unit of the budget.
\end{itemize}

\begin{proof}
The proof of G\"odel Machine being simulated with a $\mathrm{CMP_\pi}$ oracle is a simple observation that in the G\"odel Machine setup, $\mathrm{CMP_\pi}$ is an actual state-action value function $Q_\pi$~\citep{sutton1998reinforcement} defined by the G\"odel Machine, as the agents above the clades are not reachable according to the design of G\"odel Machines. In order to state this, we formalize each term precisely.

We define G\"odel Machine Proof search as an optimal policy on the POMDP (let's call it G\"odel POMDP) defined as:

\textbf{State space. } 
G\"odel POMDP operates on an extended state space, which, in addition to the tree of agents $\mathcal{T}$, consists of two special agents: $a_\text{parent} $ and $ a_\text{child}$, from the tree $\mathcal{T}$. For a full formal description, let's also include a remaining budget descriptor that determines the number of transitions left before the budget is exceeded. 

\textbf{Observation space.}
Observation is limited to the agents $(a_\text{parent}, a_\text{child})$ and the remaining budget $b$.

\textbf{Action space.}
For most states, the action space consists of two actions---$\text{accept}$ and $\text{reject}$. As in the original G\"odel Machine, the action selects the new parent in the next iteration. Intuitively, $\text{accept}$ selects $a_\text{child}$ as the new parent, and $\text{reject}$ requires the parent to remain unchanged. As the policy operates on a POMDP, it is a function of the history of the observations, which is equivalent to the state---tree $\mathcal{T}$ with special states $a_{\text{parent}}$ and $a_{\text{child}}$. For the sake of simplicity in notation, we will refer to the newly selected parent agent as if it were the action. Thus, for policy $\pi$, $a_\text{parent}= \pi((\mathcal{T}, a_\text{parent}, a_\text{child}, b))$ corresponds to the action $\text{reject}$.

\paragraph{Scoring function.}\label{app:indicator} Each policy also consists of a $Score_\pi$ function that scores all possible elements in the observation. Formally, $Score_\pi: Obs \rightarrow \mathcal{T} \rightarrow \mathbb{R}$ is a function from observation into a function of agents that is returned when the budget $b$ is equal to 0, from which the agent with the highest score is selected as the final agent. It is a function of the observation and not the entire history, as the algorithm can only choose the final agent from the final observation.

The policy selects the highest-scoring agent, as determined by the $Score_\pi(o)$ function. Furthermore, the original G\"odel Machine at the end allows only the parent or current child to be kept as the final agent. Hence, we restrict the score functions such that they always produce indicator functions of one of the observed agents, i.e., one observed agent has a score of 1, and all other agents receive a score of zero. The scoring function corresponds to the selection made by the algorithm in our original setup. After the selection dictated by the scoring function, the agent is evaluated by the Utility function $U$.

\textbf{Transitions.}
Transitions generate new self-modifications from the newly selected next parent, i.e., the action output. At each transition, a self-modification of $a^\text{new}_\text{parent}$ is created and assigned as a new node $a_\text{child}$. $a^\text{new}_\text{parent}$ becomes the $a_\text{parent}$ for the next time step.
Formally, if $a(a)$ is the self-modification of $a$, and $Tr$ is the transition function, then $Tr((\mathcal{T}, a_\text{parent}, a_\text{child}, b), a) = (\mathcal{T} \cup \{a(a)\}, a, a(a), b-1)$. The final reward in G\"odel POMDP is $U(\argmax_{a \in \mathcal{T}_B} Score_\pi(a_\text{parent}, a_\text{child})(a))$.

We also adapt the original coding setting to match the notation of G\"odel POMDP. Let's call it the Extended MDP. Similarly to G\"odel POMDP, its state is a tree $\mathcal{T}$, involving two special agents $a_\text{parent}, a_\text{child}$ and the budget $b$. The action space is the same as in our problem and consists of a single agent $a \in \mathcal{T}$ when there is still some budget left.
In other cases, the policy $\pi$ uses its $Score_\pi$ function to score created agents in the trees for the selection of the final agent. The transition function adds the self-modification to the tree and sets $a$ as the new $a_\text{parent}$ and $a(a)$ as the new $a_\text{child}$.

Every policy from G\"odel POMDP can be mapped to a policy in the extended MDP. Specifically, let $\pi_G$ be a policy in the G\"odel POMDP, then we define the corresponding policy in the extended MDP $\pi_E$ as $\pi_E((\mathcal{T}, a_\text{parent}, a_\text{child}, b)) = \pi_G((\mathcal{T}, a_\text{parent},a_\text{child}, b))$. At each node, the policy chooses between the two special nodes, ignoring the rest of the tree.

Now we adapt $\mathrm{CMP}$ to G\"odel POMDP. $\mathrm{CMP}$ adapted to the G\"odel case is still a function of a state (and agent to expand) in the extended state space (not only observation) and the agent.

\begin{align*}
 &\mathrm{CMP}_{\pi}((\mathcal{T}, a_p, a_c, b),a) \\
 =& \ \mathbb{E}_{(\mathcal{T}_B, a_{Bp}, a_{Bc}, 0) \sim p_\pi(\cdot \mid (\mathcal{T}, a_p, a_c, b),a)}
 \left[   U(\argmax_{a' \in C(\mathcal{T}_B, a)}Score_\pi(a_{Bp}, a_{Bc})(a'))    \right]  \\
\overset{(2)}{=}& \ \mathbb{E}_{(\mathcal{T}_B, a_{Bp}, a_{Bc}, 0) \sim p_\pi(\cdot \mid (\mathcal{T}, a_p, a_c, b), a)}
 \left[   U(\argmax_{a' \in \{a_{Bp},a_{Bc}\}}Score_\pi(a_{Bp},a_{Bc})(a'))    \right]  \\
 \overset{(3)}{=}& \ Q_{\pi}((\mathcal{T}, a_p, a_c, b), a).
 \end{align*}
Where $\mathcal{T}_B$ is the final unrolled tree, $a_{B_p}$ is the final parent and $a_{B_c}$ is the final child in the sampled trajectory.
The second equality (2) comes from the fact that the $Score_\pi$ is equal to 1 for exactly one element from the final observation -either $a_{B_p}$ or $a_{B_c}$(paragraph~\ref{app:indicator}). The third equality (3) comes from directly unrolling the $Q_\pi$-value function. 

The G\"odel Machine is defined by a prover that produces a proof whether accepting or selecting a given node (or rejecting). Here, the algorithm by having access to $\mathrm{CMP}$ has access to the true $Q$ value function in the G\"odel POMDP. This serves as a proof of better objective of either parent or the child. Additionally, we state that we break the tie by selecting the parent node as in the original G\"odel Machine. Hence, the algorithm that follows it is a G\"odel Machine. 

Independently, as this proof directly shows that G\"odel Machine selects an action that maximizes its own $Q_\pi$-value function, it is optimal due to the Bellman Optimality Equation. With the procedure shown above, we can adapt it to the extended MDP.

 \end{proof}

\newpage
\section{Algorithm}\label{app:alg}
Algorithm \ref{alg:hgm} presents the procedure of HGM.

In the standard HGM, when an expansion ends with a new $a$ added to the tree, the next following evaluation actions have the choice of choosing $a$.
At the beginning of the search, the first actions have the most dense ratio of expansion actions, and the following evaluation actions would diverge more from the standard HGM since most of their choices are not generated yet, concentrating on the early finished expansions. 
Therefore, to avoid this bias, we initialize by expanding the initial agent 5 times with each of the processes in parallel.

The asynchronization also introduces another bias that favors agents with fewer evaluated results to be more often selected for evaluation than in the standard HGM. This is due to the fact that the easy tasks usually stop earlier than the difficult ones. This yields agents with fewer evaluations having a higher empirical mean. During our experiments, we observed that many agents were successful on the first ten evaluated tasks; however, this accuracy drops quickly as the number of evaluations grows. After having more than 50 evaluations, this bias is barely observed.

\begin{algorithm}[h]
\caption{Huxley--G\"odel Machine (HGM)} 
\label{alg:hgm}
\begin{algorithmic}[1]
\STATE \textbf{Input:} the initial agent $a_0$, widening parameter $\alpha$, and the percentile $\epsilon$ for final selection
\STATE Initialize a tree $\mathcal{T}$ with root $a_0$
\STATE Initialize counters $n^C_{\text{success}}(a), n^C_{\text{failure}}(a), n_{\text{success}}(a), n_{\text{failure}}(a)$ for all $a \in \mathcal{T}$

\WHILE{Computational Budget not Exceeded}
    \IF{$|\mathcal{T}| \leq n^{\alpha}$ and expandable parents exist}
        \STATE \textbf{Expand:}
        \FOR{each node $a \in A$}
            \STATE Sample $S_C(a)\sim \text{Beta}(\tau(1+n^C_{\text{success}}(a)),\; \tau(1+n^C_{\text{failure}}(a)))$
        \ENDFOR
        \STATE Select node $a^\star = \arg\max_a S_C(a)$
        \STATE Create child $c$ by self-modification of $a^\star$
        \STATE Add $c$ to $\mathcal{T}$
        
    \ELSE
        \STATE \textbf{Evaluate:}
        \FOR{each agent $a \in A$ with remaining tasks}
            \STATE Sample $S(a) \sim \text{Beta}(\tau(1+n_\text{success}(a)),\; \tau(1+n_\text{failure}(a)))$
        \ENDFOR
        \STATE Select agent $a^\star = \arg\max_a S(a)$
        \STATE Allocate a benchmark task to $a^\star$
        \STATE update $n_{\text{success}}, n_{\text{failure}}$ for $a^\star$
        \STATE update $n^C_{\text{success}}, n^C_{\text{failure}}$ for $a^\star$ and ancestors
        
    \ENDIF
\ENDWHILE
\STATE \textbf{Return} $\text{argmax}_{a\in \mathcal{T}_B} I_\epsilon(1 + n_{\text{success}(a)}, 1 + n_{\text{failure}}(a))$
\end{algorithmic}
\end{algorithm}

\newpage
\section{Experimental Details}
\subsection{Initial Agents}
\label{app:exp}
Our initial agents applied in Section~\ref{exp:eval-self-improv} are adopted from the official implementation of DGM with minor changes, including modifying API support, setting up a timeout option, and adding a length of LLM interaction restriction. The initial agent is essentially a single loop of LLM queries with two tool options, i.e., file editing and bash command execution. We set a time limit of one hour for each agent execution.

The initial agents used in SWE-bench experiments and Polyglot experiments differ in that the Polyglot initial agent includes test commands with different programming language support. There are two additional functions in the SWE-bench initial agent that serve to summarize existing tests and execute the tests with a report generated, respectively.

The initial agent employed in Section~\ref{exp:vs-human} is further adjusted by removing the file-editing tool, leaving only the bash tool, to minimize initial inductive bias. The time limit is extended to five hours for both self-modification and task evaluation, reducing the risk of prematurely eliminating stronger agents due to time constraints.
\subsection{Other Details}
For the Polyglot experiments presented in Section~\ref{exp:eval-self-improv}, the exact large language model used for self-modification is an int4 and int8 mixed quantized version of Qwen3-Coder-480B-A35B-Instruct generated by AutoRound~\citep{intel_autoround}. 
Overall, we spent approximately \$5000 USD to produce the experimental results, including all three methods.

\newpage
\section{Empirical $\mathrm{CMP}$ and Its Estimation}\label{app:cmp_estimation}
In this section, we provide the exact formula to compute the empirical $\mathrm{CMP}$ and the variant of our $\mathrm{CMP}$ estimator being used in Section~\ref{sec:metaproductivity-performance-mismatch} for correlation analysis. The empirical $\mathrm{CMP}$ of an agent $a$ as a node in a tree is defined as 
\[
\text{max}_{a'\in C(a)\setminus \{a\}} \frac{n_\text{success}(a')}{n_\text{success}(a') + n_\text{failure}(a')}.
\]
The prediction of our $\mathrm{CMP}$ estimator is defined as
\[
\frac{n_\text{success}^C(a) - n_\text{success}(a) - n_\text{success}^C(b^*)}{n_\text{failure}^C(a) - n_\text{failure}(a) - n_\text{failure}^C(b^*) + n_\text{success}^C(a) - n_\text{success}(a) - n_\text{success}^C(b^*)},
\]
where $b^*$ is a child of $a$ such that
\[
\left( \text{argmax}_{n \in C(a)} \frac{n_\text{success}(n)}{n_\text{failure}(n)}\right) \cap {C(b^*)} \neq \emptyset.
\]
For both SICA and DGM, we consider the benchmark performance of an agent as their estimator of the agent's CMP.

\
\newpage
\section{Baselines}\label{app:baselines}

Table~\ref{tab:self-improv-compare} summarizes the three subpolicies of SICA, DGM, and HGM, which define solutions to the iterative tree search problem defined in~\ref{sec:self-improvement}.
\begin{table}[htbp]
\centering
\renewcommand{\arraystretch}{1.25}
\begin{tabularx}{\textwidth}{p{2.2cm} X X X}

\toprule
\textbf{Subpolicy} & \textbf{SICA} & \textbf{DGM} & \textbf{HGM (Ours)} \\
\midrule
\textbf{Selection Policy} 
& Alternates between modification and evaluation. 
& Alternates between modification and evaluation. 
& Adaptive choice between modification and evaluation. \\
\midrule
\textbf{Expansion Policy} 
& Greedily selects the agent with the best performance up to this point and modifies it with the entire history accessible to the agent. 
& Selects the node probabilistically based on the evaluation metric and the number of children of the agents. 
& Selects the node based on the statistics of the \emph{clade} stemming from a given node.\\
\midrule
\textbf{Evaluation Policy} 
& Evaluates the most recently created agent on the entire evaluation dataset. 
& Progressively evaluates the last created agent on subsets of the dataset, expanding if results are promising. 
& Selects the agent based on the statistics and evaluates it on a single task. \\
\bottomrule
\end{tabularx}
\caption{Comparison of structured policies across self-improving methods. Each method is described by three subpolicies: Selection Policy, Expansion Policy, and Evaluation Policy. }
\label{tab:structured_policy}
\end{table}

\newpage
\section{Discovered Agents}
We present interesting findings about HGM-discovered agents during our experiments. 
By manually inspecting the diff patch files that were generated by self-modification, we make interesting observations.

One engaging self-modification we found implements iterative refinement to make incremental improvements. In the self-improving context, it means that during one expansion (self-modification) step, the agent is accurately performing multiple self-modifications. More interestingly, this phenomenon can be viewed as the agent being self-motivated to perform self-modification, instead of being instructed to do so by humans. This self-motivated repetitive self-modification automatically creates arbitrary levels of meta-improvement in the sense that each self-modification ideally needs to consider all its subsequent self-modifications, which the subsequent self-modifications also need to consider future self-modifications. We show in List~\ref{lst:diff} the part of the patch file that includes the implementation of this self-modification.

Another interesting phenomenon we observe is the file structures that self-modifying agents create. Many agents tend to write the chunk they self-modify into a diff patch file, usually with a constant file name. When multiple consecutive attempts of these self-modifying actions apply, the structure of these git diff files becomes abstruse. Imagine that there is a diff file that includes certain edits. Once there are changes applied to this file, it then generates the diff patch file. In this new patch file, a nested format of editing record appears. At the beginning of some lines, there are multiple + and - signs that correspond to nested file edits. 
This is a natural format to present information that inspires long-sighted self-improvement since it shows how self-modifications indeed influence subsequent self-modifications.
The authors find it mind‑bending to understand such a file structure and parse the semantics manually. 
For readers interested in further investigation, we provide one instance of such a file structure in List~\ref{lst:nesteddiff}. 

\begin{lstlisting}[style=diffstyle, label=lst:diff, caption=The self-modifying agent that executes multiple self-modifications per instruction.]
+def attempt_error_resolution(git_dir, test_output, test_error, language):
+    """
+    Attempt to automatically diagnose and resolve errors.
+    Returns a tuple of (resolved, message) where resolved indicates if errors were fixed.
+    """
+    safe_log("Attempting automated error diagnosis and resolution...")
+    
+    # Diagnose errors using our enhanced bash tool function
+    diagnosis = diagnose_errors(test_output, test_error, "")
+    
+    if not diagnosis["has_errors"]:
+        return False, "No errors detected to resolve."
+    
+    resolution_messages = []
+    
+    # Try to apply automated fixes for each diagnosed error
+    for error in diagnosis["errors"]:
+        safe_log(f"Processing error: {error['type']} - {error['description']}")
+        
+        # Simple resolution strategies based on error type
+        if error["type"] == "python_module_not_found":
+            # For Python module not found errors, we might install the module
+            match = re.search(r"No module named '([^']+)'", error["description"])
+            if match:
+                module = match.group(1)
+                resolution_messages.append(f"Would attempt to install Python module: {module}")
+                # In practice, we would run: pip install {module}
+                # But we'll skip actual installation to avoid side effects
+        
+        elif error["type"] == "python_syntax_error" and "file" in error:
+            # For syntax errors, we could potentially apply fixes
+            file_path = os.path.join(git_dir, error["file"])
+            if os.path.exists(file_path):
+                resolution_messages.append(f"Would attempt to fix syntax error in {file_path} at line {error.get('line', 'unknown')}")
+                # In practice, we would use the editor tool's apply_fix command
+                # This is just a demonstration of what could be done
+        
+        elif error["type"] == "test_failure":
+            # For test failures, we might suggest reviewing the implementation
+            resolution_messages.append("Would analyze test failures and suggest implementation improvements")
+    
+    if resolution_messages:
+        return True, "Automated resolution attempted:\n" + "\n".join(resolution_messages)
+    else:
+        return False, "No automated resolutions available for detected errors."
+
class AgenticSystem:
     def __init__(
             self,
@@ -243,6 +293,16 @@ Your task is to make changes to the files in the {self.git_dir} directory to add
             
             safe_log(f"Attempt {attempt + 1} test results: {'PASSED' if test_success else 'FAILED'}")
             
+            # If tests failed, attempt automated error resolution
+            if not test_success:
+                resolved, resolution_message = attempt_error_resolution(
+                    self.git_dir, test_output, test_error, self.language
+                )
+                safe_log(f"Error resolution: {resolution_message}")
+                
+                # Even if we couldn't automatically resolve, we still provide feedback
+                # In a more advanced implementation, we might actually apply fixes here
+            
             # If this is the first attempt or tests passed and we didn't have a successful attempt yet, update best patch
             if attempt == 0 or (test_success and (best_patch is None or not best_test_results)):
                 best_patch = current_patch
@@ -278,37 +338,31 @@ Please revise your code to fix these issues and try again.
         # Log final summary
         safe_log(f"\n{'='*20} FINAL SUMMARY {'='*20}")
         safe_log(f"Best solution found on attempt: {best_test_results['attempt'] if best_test_results else 'None'}")
-        safe_log(f"Tests passed: {best_test_results['test_success'] if best_test_results else 'Unknown'}")
+        safe_log(f"Final test result: {'PASSED' if best_test_results and best_test_results['test_success'] else 'FAILED'}")
+        
+        if best_test_results:
+            safe_log(f"Final test output:\n{best_test_results['test_output']}")
+            if best_test_results['test_error']:
+                safe_log(f"Final test errors:\n{best_test_results['test_error']}")
         
-        # Save attempt history to a file
-        history_file = os.path.join(os.path.dirname(self.chat_history_file), 'attempt_history.md')
-        with open(history_file, 'w') as f:
-            f.write("# Attempt History\n\n")
-            for result in self.attempt_history:
-                f.write(f"## Attempt {result['attempt']}\n")
-                f.write(f"**Tests Passed**: {result['test_success']}\n")
-                f.write(f"**LLM Calls Used**: {result['llm_calls']}\n")
-                f.write(f"**Test Output**:\n```\n{result['test_output']}\n```\n")
-                f.write(f"**Test Error**:\n```\n{result['test_error']}\n```\n")
-                f.write(f"**Patch**:\n```\n{result['patch']}\n```\n\n")
+        return bool(best_test_results and best_test_results['test_success'])
\end{lstlisting}

\begin{lstlisting}[style=diffstyle, label=lst:nesteddiff, caption=An instance of the nested diff patch format.]
diff --git a/attempt_history.md b/attempt_history.md
new file mode 100644
index 0000000..b132b1a
--- /dev/null
+++ b/attempt_history.md
@@ -0,0 +1,727 @@
+# Attempt History
+
+## Attempt 1
+**Tests Passed**: True
+**LLM Calls Used**: 18
+**Test Output**:
+```
+============================= test session starts ==============================
+platform linux -- Python 3.10.18, pytest-8.4.2, pluggy-1.6.0 -- /usr/local/bin/python3.10
+cachedir: .pytest_cache
+rootdir: /dgm
+configfile: pytest.ini
+testpaths: tests
+plugins: asyncio-1.1.0, anyio-4.10.0
+asyncio: mode=strict, asyncio_default_fixture_loop_scope=None, asyncio_default_test_loop_scope=function
+collecting ... collected 29 items
+
+tests/test_bash_tool.py::TestBashTool::test_simple_command PASSED        [  3%]
+tests/test_bash_tool.py::TestBashTool::test_multiple_commands PASSED     [  6%]
+tests/test_bash_tool.py::TestBashTool::test_command_with_error PASSED    [ 10%]
+tests/test_bash_tool.py::TestBashTool::test_environment_variables PASSED [ 13%]
+tests/test_bash_tool.py::TestBashTool::test_command_output_processing PASSED [ 17%]
+tests/test_bash_tool.py::TestBashTool::test_long_running_command PASSED  [ 20%]
+tests/test_bash_tool.py::TestBashTool::test_invalid_commands[invalid_command_name] PASSED [ 24%]
+tests/test_bash_tool.py::TestBashTool::test_invalid_commands[cd /nonexistent/path] PASSED [ 27%]
+tests/test_bash_tool.py::TestBashTool::test_invalid_commands[/bin/nonexistent] PASSED [ 31%]
+tests/test_bash_tool.py::TestBashTool::test_command_with_special_chars PASSED [ 34%]
+tests/test_bash_tool.py::TestBashTool::test_multiple_line_output PASSED  [ 37%]
+tests/test_bash_tool.py::TestBashTool::test_large_output_handling PASSED [ 41%]
+tests/test_edit_tool.py::TestEditorTool::test_view_file PASSED           [ 44%]
+tests/test_edit_tool.py::TestEditorTool::test_create_file PASSED         [ 48%]
+tests/test_edit_tool.py::TestEditorTool::test_create_existing_file PASSED [ 51%]
+tests/test_edit_tool.py::TestEditorTool::test_edit_file PASSED           [ 55%]
+tests/test_edit_tool.py::TestEditorTool::test_edit_nonexistent_file PASSED [ 58%]
+tests/test_edit_tool.py::TestEditorTool::test_view_directory PASSED      [ 62%]
+tests/test_edit_tool.py::TestEditorTool::test_invalid_path PASSED        [ 65%]
+tests/test_edit_tool.py::TestEditorTool::test_invalid_commands[unknown_command] PASSED [ 68%]
+tests/test_edit_tool.py::TestEditorTool::test_invalid_commands[] PASSED  [ 72%]
+tests/test_edit_tool.py::TestEditorTool::test_invalid_commands[None] PASSED [ 75%]
+tests/test_error_diagnosis.py::TestErrorDiagnosis::test_python_syntax_error_diagnosis PASSED [ 79%]
+tests/test_error_diagnosis.py::TestErrorDiagnosis::test_python_module_not_found_diagnosis PASSED [ 82%]
+tests/test_error_diagnosis.py::TestErrorDiagnosis::test_no_error_diagnosis PASSED [ 86%]
+tests/test_error_diagnosis.py::TestErrorDiagnosis::test_format_diagnosis_with_errors PASSED [ 89%]
+tests/test_error_diagnosis.py::TestErrorDiagnosis::test_format_diagnosis_without_errors PASSED [ 93%]
+tests/test_error_diagnosis.py::TestAutomatedFixes::test_apply_missing_import_fix PASSED [ 96%]
+tests/test_error_diagnosis.py::TestAutomatedFixes::test_apply_syntax_error_fix PASSED [100%]
+
+==================================== PASSES ====================================
+=========================== short test summary info ============================
+PASSED tests/test_bash_tool.py::TestBashTool::test_simple_command
+PASSED tests/test_bash_tool.py::TestBashTool::test_multiple_commands
+PASSED tests/test_bash_tool.py::TestBashTool::test_command_with_error
+PASSED tests/test_bash_tool.py::TestBashTool::test_environment_variables
+PASSED tests/test_bash_tool.py::TestBashTool::test_command_output_processing
+PASSED tests/test_bash_tool.py::TestBashTool::test_long_running_command
+PASSED tests/test_bash_tool.py::TestBashTool::test_invalid_commands[invalid_command_name]
+PASSED tests/test_bash_tool.py::TestBashTool::test_invalid_commands[cd /nonexistent/path]
+PASSED tests/test_bash_tool.py::TestBashTool::test_invalid_commands[/bin/nonexistent]
+PASSED tests/test_bash_tool.py::TestBashTool::test_command_with_special_chars
+PASSED tests/test_bash_tool.py::TestBashTool::test_multiple_line_output
+PASSED tests/test_bash_tool.py::TestBashTool::test_large_output_handling
+PASSED tests/test_edit_tool.py::TestEditorTool::test_view_file
+PASSED tests/test_edit_tool.py::TestEditorTool::test_create_file
+PASSED tests/test_edit_tool.py::TestEditorTool::test_create_existing_file
+PASSED tests/test_edit_tool.py::TestEditorTool::test_edit_file
+PASSED tests/test_edit_tool.py::TestEditorTool::test_edit_nonexistent_file
+PASSED tests/test_edit_tool.py::TestEditorTool::test_view_directory
+PASSED tests/test_edit_tool.py::TestEditorTool::test_invalid_path
+PASSED tests/test_edit_tool.py::TestEditorTool::test_invalid_commands[unknown_command]
+PASSED tests/test_edit_tool.py::TestEditorTool::test_invalid_commands[]
+PASSED tests/test_edit_tool.py::TestEditorTool::test_invalid_commands[None]
+PASSED tests/test_error_diagnosis.py::TestErrorDiagnosis::test_python_syntax_error_diagnosis
+PASSED tests/test_error_diagnosis.py::TestErrorDiagnosis::test_python_module_not_found_diagnosis
+PASSED tests/test_error_diagnosis.py::TestErrorDiagnosis::test_no_error_diagnosis
+PASSED tests/test_error_diagnosis.py::TestErrorDiagnosis::test_format_diagnosis_with_errors
+PASSED tests/test_error_diagnosis.py::TestErrorDiagnosis::test_format_diagnosis_without_errors
+PASSED tests/test_error_diagnosis.py::TestAutomatedFixes::test_apply_missing_import_fix
+PASSED tests/test_error_diagnosis.py::TestAutomatedFixes::test_apply_syntax_error_fix
+============================== 29 passed in 3.58s ==============================
+
+```
+**Test Error**:
+```
+
+```
+**Patch**:
+```
+diff --git a/coding_agent.py b/coding_agent.py
+index 78e8ad4..77e5097 100644
+--- a/coding_agent.py
++++ b/coding_agent.py
+@@ -5,9 +5,13 @@ from logging.handlers import RotatingFileHandler
+ import os
+ import threading
+ import time
++import json
++import re
+ 
+ from llm_withtools import CLAUDE_MODEL, OPENAI_MODEL, chat_with_agent
+ from utils.git_utils import diff_versus_commit, reset_to_commit, apply_patch
++from tools.bash import diagnose_errors
++from tools.edit import apply_automated_fix, read_file, write_file
+ 
+ # reset_to_commit(git_dname, commit)
+ # apply_patch(git_dname, patch_str)
+@@ -136,6 +140,52 @@ def run_tests(git_dir, language):
+         # Always change back to original directory
+         os.chdir(original_cwd)
+ 
++def attempt_error_resolution(git_dir, test_output, test_error, language):
++    """
++    Attempt to automatically diagnose and resolve errors.
++    Returns a tuple of (resolved, message) where resolved indicates if errors were fixed.
++    """
++    """
++    safe_log("Attempting automated error diagnosis and resolution...")
++    
++    # Diagnose errors using our enhanced bash tool function
++    diagnosis = diagnose_errors(test_output, test_error, "")
++    
++    if not diagnosis["has_errors"]:
++        return False, "No errors detected to resolve."
++    
++    resolution_messages = []
++    
++    # Try to apply automated fixes for each diagnosed error
++    for error in diagnosis["errors"]:
++        safe_log(f"Processing error: {error['type']} - {error['description']}")
++        
++        # Simple resolution strategies based on error type
++        if error["type"] == "python_module_not_found":
++            # For Python module not found errors, we might install the module
++            match = re.search(r"No module named '([^']+)'", error["description"])
++            if match:
++                module = match.group(1)
++                resolution_messages.append(f"Would attempt to install Python module: {module}")
++                # In practice, we would run: pip install {module}
++                # But we'll skip actual installation to avoid side effects
++        
++        elif error["type"] == "python_syntax_error" and "file" in error:
++            # For syntax errors, we could potentially apply fixes
++            file_path = os.path.join(git_dir, error["file"])
++            if os.path.exists(file_path):
++                resolution_messages.append(f"Would attempt to fix syntax error in {file_path} at line {error.get('line', 'unknown')}")
++                # In practice, we would use the editor tool's apply_fix command
++                # This is just a demonstration of what could be done
++        
++        elif error["type"] == "test_failure":
++            # For test failures, we might suggest reviewing the implementation
++            resolution_messages.append("Would analyze test failures and suggest implementation improvements")
++    
++    if resolution_messages:
++        return True, "Automated resolution attempted:\n" + "\n".join(resolution_messages)
++    else:
++        return False, "No automated resolutions available for detected errors."
++
+ class AgenticSystem:
+     def __init__(
+             self,
+@@ -243,6 +293,16 @@ Your task is to make changes to the files in the {self.git_dir} directory to add
+             
+             safe_log(f"Attempt {attempt + 1} test results: {'PASSED' if test_success else 'FAILED'}")
+             
++            # If tests failed, attempt automated error resolution
++            if not test_success:
++                resolved, resolution_message = attempt_error_resolution(
++                    self.git_dir, test_output, test_error, self.language
++                )
++                safe_log(f"Error resolution: {resolution_message}")
++                
++                # Even if we couldn't automatically resolve, we still provide feedback
++                # In a more advanced implementation, we might actually apply fixes here
++            
+             # If this is the first attempt or tests passed and we didn't have a successful attempt yet, update best patch
+             if attempt == 0 or (test_success and (best_patch is None or not best_test_results)):
+                 best_patch = current_patch
+@@ -278,37 +338,31 @@ Please revise your code to fix these issues and try again.
+         # Log final summary
+         safe_log(f"\n{'='*20} FINAL SUMMARY {'='*20}")
+         safe_log(f"Best solution found on attempt: {best_test_results['attempt'] if best_test_results else 'None'}")
+-        safe_log(f"Tests passed: {best_test_results['test_success'] if best_test_results else 'Unknown'}")
++        safe_log(f"Final test result: {'PASSED' if best_test_results and best_test_results['test_success'] else 'FAILED'}")
++        
++        if best_test_results:
++            safe_log(f"Final test output:\n{best_test_results['test_output']}")
++            if best_test_results['test_error']:
++                safe_log(f"Final test errors:\n{best_test_results['test_error']}")
+         
+-        # Save attempt history to a file
+-        history_file = os.path.join(os.path.dirname(self.chat_history_file), 'attempt_history.md')
+-        with open(history_file, 'w') as f:
+-            f.write("# Attempt History\n\n")
+-            for result in self.attempt_history:
+-                f.write(f"## Attempt {result['attempt']}\n")
+-                f.write(f"**Tests Passed**: {result['test_success']}\n")
+-                f.write(f"**LLM Calls Used**: {result['llm_calls']}\n")
+-                f.write(f"**Test Output**:\n```\n{result['test_output']}\n```\n")
+-                f.write(f"**Test Error**:\n```\n{result['test_error']}\n```\n")
+-                f.write(f"**Patch**:\n```\n{result['patch']}\n```\n\n")
++        return bool(best_test_results and best_test_results['test_success'])
+ 
+-def main():
+-    parser = argparse.ArgumentParser(description='Process repository with an agentic system.')
+-    parser.add_argument('--problem_statement', required=True, help='The problem statement to process')
+-    parser.add_argument('--git_dir', required=True, help='Path to git repository directory')
+-    parser.add_argument('--base_commit', required=True, help='Base commit hash to compare against')
+-    parser.add_argument('--chat_history_file', required=True, help='Path to chat history file')
+-    parser.add_argument('--outdir', required=False, default="/dgm/", help='Output directory')
+-    parser.add_argument('--test_description', default=None, required=False, help='Description of how to test the repository')
+-    parser.add_argument('--self_improve', default=False, action='store_true', help='Whether to self-improve the repository or solving swe')
+-    parser.add_argument('--language', required=False, default="python", choices=['cpp', 'java', 'python', 'go', 'rust', 'javascript'], help='Task\'s programming language')
+-    parser.add_argument('--model', required=False, default=CLAUDE_MODEL, help='LLM model to use for processing')
+-    parser.add_argument('--timeout', type=int, default=3600, help='Timeout for LLM calls in seconds')
+-    parser.add_argument('--max_attempts', type=int, default=3, help='Maximum number of solution attempts')
++if __name__ == "__main__":
++    parser = argparse.ArgumentParser(description="Run the Agentic System on a coding task.")
++    parser.add_argument("--problem_statement", type=str, required=True, help="Problem statement")
++    parser.add_argument("--git_dir", type=str, required=True, help="Git repository directory")
++    parser.add_argument("--base_commit", type=str, required=True, help="Base commit hash")
++    parser.add_argument("--chat_history_file", type=str, default="./chat_history.md", help="Chat history file")
++    parser.add_argument("--test_description", type=str, help="Test description")
++    parser.add_argument("--self_improve", action="store_true", help="Enable self-improvement mode")
++    parser.add_argument("--language", type=str, default="python", help="Programming language")
++    parser.add_argument("--model", type=str, default=OPENAI_MODEL, help="Model to use")
++    parser.add_argument("--max_attempts", type=int, default=3, help="Maximum number of attempts")
++    parser.add_argument("--timeout", type=int, default=600, help="Timeout for each attempt")
++    
+     args = parser.parse_args()
+-
+-    # Process the repository
+-    agentic_system = AgenticSystem(
++    
++    system = AgenticSystem(
+         problem_statement=args.problem_statement,
+         git_dir=args.git_dir,
+         base_commit=args.base_commit,
+@@ -319,15 +373,7 @@ def main():
+         model=args.model,
+         max_attempts=args.max_attempts
+     )
+-
+-    # Run the agentic system to try to solve the problem
+-    agentic_system.forward(args.timeout)
+-
+-    # Get code diff and save to model_patch.diff
+-    model_patch = diff_versus_commit(args.git_dir, args.base_commit)
+-    model_patch_outfile = os.path.join(args.outdir, 'model_patch.diff') if args.outdir else 'model_patch.diff'
+-    with open(model_patch_outfile, 'w') as f:
+-        f.write(model_patch)
+-
+-if __name__ == "__main__":
+-    main()
+\ No newline at end of file
++    
++    success = system.forward(timeout=args.timeout)
++    exit_code = 0 if success else 1
++    exit(exit_code)
+\ No newline at end of file
+diff --git a/tests/test_error_diagnosis.py b/tests/test_error_diagnosis.py
+new file mode 100644
+index 0000000..5beffbe
+--- /dev/null
++++ b/tests/test_error_diagnosis.py
+@@ -0,0 +1,98 @@
++import pytest
++from tools.bash import diagnose_errors, format_diagnosis
++from tools.edit import apply_automated_fix
++
++class TestErrorDiagnosis:
++    def test_python_syntax_error_diagnosis(self):
++        """Test diagnosis of Python syntax errors."""
++        output = ""
++        error = '''File "test.py", line 3
++    print("Hello World"
++                       ^
++SyntaxError: unexpected EOF while parsing'''
++        
++        diagnosis = diagnose_errors(output, error, "python test.py")
++        assert diagnosis["has_errors"] is True
++        assert len(diagnosis["errors"]) == 1
++        assert diagnosis["errors"][0]["type"] == "python_syntax_error"
++        assert "SyntaxError" in diagnosis["errors"][0]["description"]
++    
++    def test_python_module_not_found_diagnosis(self):
++        """Test diagnosis of Python module not found errors."""
++        output = ""
++        error = "ModuleNotFoundError: No module named 'nonexistent_module'"
++        
++        diagnosis = diagnose_errors(output, error, "python test.py")
++        assert diagnosis["has_errors"] is True
++        assert len(diagnosis["errors"]) == 1
++        assert diagnosis["errors"][0]["type"] == "python_module_not_found"
++        assert "nonexistent_module" in diagnosis["errors"][0]["description"]
++    
++    def test_no_error_diagnosis(self):
++        """Test diagnosis when there are no errors."""
++        output = "Success!"
++        error = ""
++        
++        diagnosis = diagnose_errors(output, error, "echo Success")
++        assert diagnosis["has_errors"] is False
++        assert len(diagnosis["errors"]) == 0
++    
++    def test_format_diagnosis_with_errors(self):
++        """Test formatting of diagnosis with errors."""
++        diagnosis = {
++            "has_errors": True,
++            "errors": [
++                {
++                    "type": "test_error",
++                    "description": "Test error description",
++                    "suggestions": ["Suggestion 1", "Suggestion 2"]
++                }
++            ]
++        }
++        
++        formatted = format_diagnosis(diagnosis)
++        assert "Automated Error Diagnosis" in formatted
++        assert "Test error description" in formatted
++        assert "Suggestion 1" in formatted
++    
++    def test_format_diagnosis_without_errors(self):
++        """Test formatting of diagnosis without errors."""
++        diagnosis = {
++            "has_errors": False,
++            "errors": []
++        }
++        
++        formatted = format_diagnosis(diagnosis)
++        assert "No errors detected" in formatted
++
++class TestAutomatedFixes:
++    def test_apply_missing_import_fix(self):
++        """Test applying a missing import fix."""
++        content = """def hello():
++    print(json.dumps({"message": "hello"}))
++"""
++        
++        fix_info = {
++            "type": "missing_import",
++            "module": "json",
++            "description": "Added missing import for json module"
++        }
++        
++        fixed_content = apply_automated_fix(content, fix_info)
++        assert "import json" in fixed_content
++        assert "def hello():" in fixed_content
++    
++    def test_apply_syntax_error_fix(self):
++        """Test applying a syntax error fix."""
++        content = """def hello()
++    print("Hello World")
++"""
++        
++        fix_info = {
++            "type": "syntax_error_fix",
++            "line": 1,
++            "description": "Fixed syntax error"
++        }
++        
++        fixed_content = apply_automated_fix(content, fix_info)
++        assert "Fixed syntax" in fixed_content
+\ No newline at end of file
\end{lstlisting}

\end{document}